\pdfoutput=1

\documentclass[11pt]{article}

\usepackage[final]{acl}
\usepackage{amsmath}
\usepackage{amsfonts,amssymb}
\usepackage{times}
\usepackage{CJK}
\usepackage{CJK}
\usepackage{latexsym}
\usepackage{latexsym}
\usepackage{subfiles}
\usepackage{booktabs}
\usepackage{multirow}
\usepackage{array,tabularx,caption,boldline}
\usepackage{arydshln}
\usepackage[T1]{fontenc}

\usepackage[utf8]{inputenc}

\usepackage{microtype}
\usepackage{ragged2e}
\usepackage{inconsolata}
\usepackage{graphicx}
\usepackage{algorithm}
\usepackage{algorithmic}
\usepackage{tabularray}
\usepackage{makecell}
\usepackage{makebox}
\newcommand{\ignore}[1]{}

\title{DeMPT: Decoding-enhanced Multi-phase Prompt Tuning for Making LLMs Be Better Context-aware Translators}

\author{Xinglin Lyu$^{\clubsuit}$, ~Junhui Li$^{\clubsuit}$\thanks{ Corresponding author: Junhui Li.}, ~Yanqing Zhao$^{\spadesuit}$, ~Min Zhang$^{\spadesuit}$, ~Daimeng Wei$^{\spadesuit}$, \\ {\bf Shimin Tao$^{\spadesuit}$, ~Hao Yang$^{\spadesuit}$,} {\bf ~Min Zhang$^{\clubsuit}$} \\
  $^{\clubsuit}$School of Computer Science and Technology, Soochow University, Suzhou, China \\
  $^{\spadesuit}$Huawei Translation Services Center, Beijing, China \\
  \texttt{xllv2020@stu.suda.edu.cn}, \texttt{\{lijunhui,minzhang\}@suda.edu.cn}\\
\texttt{\{zhaoyanqing,zhangmin186,weidaimeng,taoshimin,yanghao30\}@huawei.com}
}

\begin{document}
\maketitle
\begin{abstract}
Generally, the {\it decoder-only} large language models (LLMs) are adapted to context-aware neural machine translation (NMT) in a concatenating way, where LLMs take the concatenation of the source sentence (i.e., intra-sentence context) and the inter-sentence context as the input, and then to generate the target tokens sequentially. This adaptation strategy, i.e., concatenation mode, considers intra-sentence and inter-sentence contexts with the same priority, despite an apparent difference between the two kinds of contexts. In this paper, we propose an alternative adaptation approach, named {\bf D}ecoding-{\bf e}nhanced {\bf M}ulti-phase {\bf P}rompt {\bf T}uning (DeMPT), to make LLMs discriminately model and utilize the inter- and intra-sentence context and more effectively adapt LLMs to context-aware NMT. First, DeMPT divides the context-aware NMT process into three separate phases. During each phase, different continuous prompts are introduced to make LLMs discriminately model various information. Second, DeMPT employs a heuristic way 
to further discriminately enhance the utilization of the source-side inter- and intra-sentence information at the final decoding phase. Experiments show that our approach significantly outperforms the concatenation method, and further improves the performance of LLMs in discourse modeling.\footnote{We release our code and datasets on \url{https://github.com/xllyu-nlp/DeMPT}.}
\end{abstract}
\section{Introduction}
\begin{figure*}
    \centering
    \resizebox{0.9\textwidth}{!}{
    \includegraphics{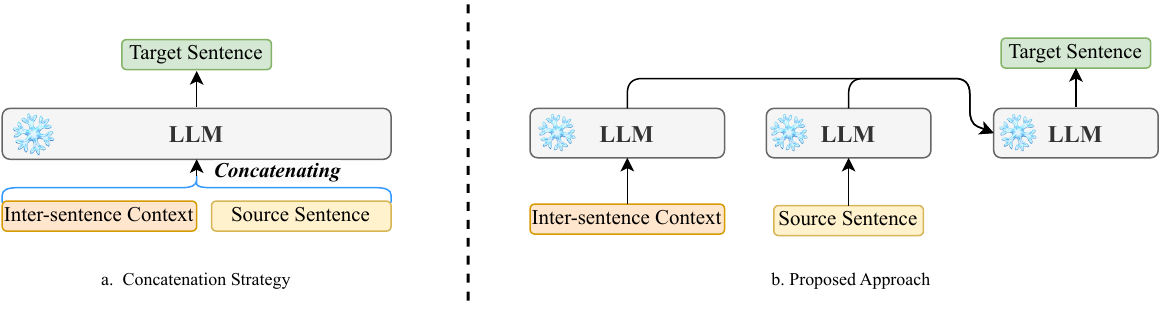}}
    \caption{Comparison of different strategies for adapting LLMs to context-aware NMT. The concatenation strategy (\textit{left}) treats inter-sentence and intra-sentence (referred to as the "source sentence" context in the figure) with equal importance. In contrast, our approach (\textit{right}) divides context-aware NMT into three distinct phases, enabling LLMs to selectively model and leverage both inter- and intra-sentence contexts.}
    \label{fig:concat-phase}
\end{figure*}

Context-aware neural machine translation (NMT) goes beyond sentence-level NMT by incorporating inter-sentence context at the document level~\cite{zhang_etal_emnlp_2018,miculicich_etal_emnlp_2018,voita_etal_acl_2018,voita_etal_acl_2019,voita_etal_emnlp_2019,bao_etal_acl_2021_gtrans,sun_etal_2022_acl_rethinking}, aiming to address discourse-related challenges such as zero pronoun translation~\cite{wang-etal-2019-one}, lexical translation consistency~\cite{lyu_etal_2021_emnlp_ltcr,lyu_etal_2022_emnlp_lexicachain}, and discourse structure~\cite{hu-wan-2023-exploring}. A recent paradigm shift has been witnessed in context-aware NMT with the emergence of the \textit{decoder-only} large language models (LLMs)~\cite{Scao2022BLOOMA1,Chowdhery2022PaLMSL,Touvron2023LLaMAOA,Touvron2023Llama2O,OpenAI2023GPT4TR}. These generative language models, trained on {massive data}, have gained significant attention in the natural language processing (NLP) community. In adapting LLMs to context-aware NMT, a common strategy involves concatenating multiple source sentences as a prefix and generating translations token-by-token, relying on the prefix and previously predicted target tokens, as shown in Figure~\ref{fig:concat-phase} (a). However, a critical observation of this strategy reveals a potential drawback – the equal prioritization of the inter- and intra-sentence contexts during token generation. Importantly, the intra-sentence context inherently contains richer parallel semantic information with the target sentence and should be given a higher priority than the inter-sentence context. Consequently, we propose that separately modeling and utilizing the inter- and intra-sentence contexts should explicitly inform LLMs of the document-level context and the current sentence itself, thus being able to prevent the misallocation of attention weights to source-side tokens~\cite{bao_etal_acl_2021_gtrans,li-etal-taslp-2023-ptransfomer}. Inspired by the success of prompt tuning~\cite{li-liang-2021-prefix,liu2022pt-v2, tan-etal-2022-msp}, our alternative approach, named Decoding-Enhanced Multi-phase Prompt Tuning (DeMPT), aims to enhance LLMs' adaptability to context-aware NMT, as shown in Figure~\ref{fig:concat-phase} (b).\footnote{{Following the findings of \citet{bao_etal_acl_2021_gtrans}, which indicate that source-side context is relatively more important for document-level MT compared to target-side context, we focus exclusively on source-side context in this paper. Nonetheless, we provide an additional discussion on integrating target-side context in Appendix \ref{apdx:tgt-ctx}.}}

Specifically, we divide the whole procedure of context-aware NMT into three phases: inter-sentence context encoding, intra-sentence context encoding, and decoding. Following~\citet{li-liang-2021-prefix,liu2022pt-v2}, we sequentially and differentially adapt LLMs for each phase, utilizing phase-specific trainable prompts. This phased tuning method enables LLMs to independently capture and model both inter- and intra-sentence contexts, facilitating a better understanding of their differences. Our approach splits the input into three parts without significantly increasing computational load, thus maintaining inference speed comparable to concatenation, as detailed in Appendix \ref{apdx:speed}.

Furthermore, during the decoding phase, we propose a heuristic method to emphasize the difference between inter- and intra-sentence contexts, and avoid \textit{long-distance} issue when utilizing inter-sentence context. Specifically, at each decoding step, we use LLMs to predict the next token three times. The decoding states used for each prediction directly concatenate with the representations of two contexts in a discriminative manner. Finally, we combine three probability distributions to search for the next token as the output from the target vocabulary. This method enables LLMs to learn not only to properly capture inter-sentence context in addressing discourse-related issues but also to recognize a difference between inter- and intra-sentence contexts, allowing for effective utilization of both types of contexts.

Our contributions can be summarized as follows:
\begin{itemize}
    \item We introduce a multi-phase prompt tuning approach that divides context-aware NMT into three phases, enabling LLMs to distinguish between inter- and intra-sentence contexts.

    \item We introduce a enhanced decoding method that discriminately utilize both context types. This allows LLMs not only properly capture inter-sentence context in addressing discourse-related issues, but also be aware of the importance of the intra-sentence context.
    
    \item We validate our approach using \texttt{llama-2-7b} and \texttt{bloomz-7b1-mt} as foundation models, demonstrating its effectiveness across five translation directions. Extensive analyses further highlight the substantial enhancement in LLMs' ability for context-aware MT.
\end{itemize}
\section{Methodology}
In this section, we describe our decoding-enhanced multi-phase approach for adapting LLMs to context-aware NMT in details. Specifically, we break down the whole procedure of context-aware NMT into three phases (Section \ref{sec:ctx-enc}), i.e., inter-sentence context encoding, intra-sentence encoding, and decoding. Additionally, {we discriminatively enhance the utilization of inter- and intra-sentence contexts during the decoding phase (Section \ref{sec:cur-dec}).} Finally, we describe our phase-aware prompts and training objective in Section~\ref{sec:phase-aware} and Section~\ref{sec:train-obj}, respectively.

For a given document pair {\small $(\mathcal{S}, \mathcal{T})$} with {\small $K$} sentences, we will construct {\small $K$} training instances. Each training instance is denoted as a tuple {\small $(\mathcal{C}, S, T)$}. Here {\small$S = x|_k^{|S|}$} represents $k$-th current source sentence with {\small$|S|$} tokens, i.e., intra-sentence context, and {\small$T = y|_k^{|T|}$} is the $k$-th target sentence with {\small$|T|$} tokens. {\small $\mathcal{C}$} denotes the $z$ previous sentences of {\small$S$}, i.e., the inter-sentence context of {\small$S$}. We denote the hidden size of the LLM as $d$, and $L$ as the number of transformer layers within it.

\subsection{Multi-phase Encoding and Decoding}
\label{sec:ctx-enc}
\begin{figure}
\centering
\resizebox{\columnwidth}{!}{
\includegraphics[width=0.9\linewidth, trim=26 0 7 0,clip]{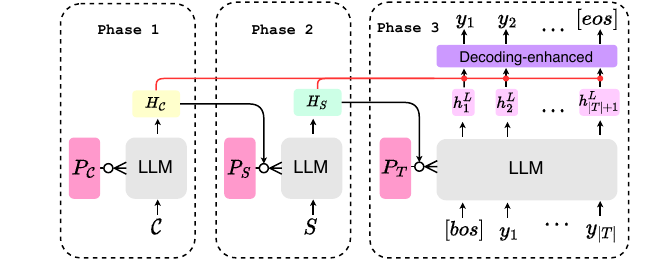}}
    \caption{Illustration of pipeline of multi-phase prompt tuning LLM for context-aware NMT. \textcolor{red}{Red} lines illustrate the procedure of enhanced decoding phase.}
    \label{fig:msp-frame}
\end{figure}

We implement our approach based on deep prompt tuning \cite{li-liang-2021-prefix, liu2022pt-v2}. Next, we use training instance {\small $(\mathcal{C}, S, T)$} as an example to describe the multi-phase approach. Figure \ref{fig:msp-frame} illustrates the procedure of multi-phase prompt tuning.

\paragraph{Inter-sentence Context Encoding Phase.} In the inter-sentence context encoding phase (Phase 1 in Figure~\ref{fig:msp-frame}), we first concatenate all sentences in {\small $\mathcal{C}$} into a sequence, and then utilize the LLM to encode {\small $\mathcal{C}$} by incorporating the trainable prompt:
\begin{equation}
\small
H_{\mathcal{C}}^{1:L} = \text{LLM}(\mathcal{C}, {\bf P}_{\mathcal{C}}),
\end{equation}
where {\small$H_{\mathcal{C}}^{1:L} \in \mathbb{R}^{L\times |\mathcal{C}|\times d}$} is the sequence of activations for {\small$\mathcal{C}$}, {\small${\bf P}_{\mathcal{C}} \in \mathbb{R}^{L \times 2q \times d}$} is the current-phase trainable prompt, and $q$ is a hyper-parameter for the length of the prompt. {\small${\bf P}_{\mathcal{C}}$} aims to adapt the LLM for better modeling the inter-sentence context. Same as basic deep prompting, at the $l$-th transformer block, we inject corresponding prompt in {\small${\bf P}_{\mathcal{C}}$} into encoding procedure of {\small$\mathcal{C}$} as follows:
\begin{equation}
\small
H_{\mathcal{C}}^l = \text{FFN}\left( \text{Multi-Attn} \left(\textbf{K}_\mathcal{C}, \textbf{V}_\mathcal{C}, \textbf{Q}_\mathcal{C}\right)\right),
\end{equation}
\begin{equation}
\small
\textbf{Q}_\mathcal{C} = H_{\mathcal{C}}^{l-1},
\end{equation}
\begin{equation}
\small
\textbf{K}_\mathcal{C} = [{\bf P}_{\mathcal{C}}[l,:q,:];H_{\mathcal{C}}^{l-1}],
\end{equation}
\begin{equation}
\small
\textbf{V}_\mathcal{C} = [{\bf P}_{\mathcal{C}}[l,q:,:];H_{\mathcal{C}}^{l-1}],
\end{equation}
where {\small$H_{\mathcal{C}}^l \in \mathbb{R}^{|\mathcal{C}|\times d}$} is the output of the $l$-th transformer block. FFN and Multi-Attn are the feed-forward network and multi-head self-attention sublayers, respectively.\footnote{For simplicity, we omit the normalization and residual operations in this paper.} $[\cdot;\cdot]$ and $[\cdot:\cdot]$ are the concatenating and slicing operations, respectively.

\paragraph{Intra-sentence Context Encoding Phase.} In the intra-sentence context encoding phase (Phase 2 in 
Figure \ref{fig:msp-frame}), the LLM encodes the intra-sentence context {\small$S$} by conditioning on the past activations of the inter-sentence context {\small$H_{\mathcal{C}}^{1:L}$} and trainable prompt:
\begin{equation}
\small
H_{S}^{1:L} = \text{LLM}(S, H_{\mathcal{C}}^{1:L}, {\bf P}_{S}), 
\end{equation}
where {\small$H_{{S}}^{1:L} \in \mathbb{R}^{L\times |S|\times d}$} is the sequence of activations for {\small$S$}, and {\small ${\bf P}_{S}\in \mathbb{R}^{L \times 2q \times d}$} denotes current-phase prompt. Similarly, at the $l$-th transformer block, we incorporate {\small $H_{{\mathcal{C}}}$} and {\small ${\bf P}_{S}$} into the encoding procedure of {\small$S$} as follows:
\begin{equation}
\small
H_{S}^l = \text{FFN}\left( \text{Multi-Attn} \left(\textbf{K}_{S}, \textbf{V}_{S}, \textbf{Q}_{S} \right)\right),
\end{equation}\begin{equation}
\small
\textbf{Q}_{S} = H_{S}^{l-1},
\end{equation}
\begin{equation}
\small
\textbf{K}_{S} = [{\bf P}_{S}[l,:q,:];H_{\mathcal{C}}^{l-1}; H_{S}^{l-1}],
\end{equation}
\begin{equation}
\small
\textbf{V}_{S} = [{\bf P}_{S}[l,q:,:];H_{\mathcal{C}}^{l-1}; H_{S}^{l-1}],
\end{equation}
where {\small $H_{S}^{l}$} is output of the $l$-th transformer block, which fuses {\small $H_{\mathcal{C}}^{l-1}$}, the $l-1$ layer output of the inter-sentence context encoding. 

\paragraph{Decoding Phase.} In the decoding phase (Phase 3 in Figure \ref{fig:msp-frame}), given the past activations {\small$H_{S}$} and trainable prompt, we call the LLM again to generate the hidden state for predicting the probability of the target sentence:
\begin{equation}
\small
H_{T}^{1:L} = \text{LLM}(T, H_{{S}}^{1:L}, {\bf P}_{T}), 
\end{equation}
where {\small$H_{T}^{1:L} \in \mathbb{R}^{L\times |T|\times d}$} is the sequence of activations for {\small$T$}, and {\small ${\bf P}_{T}\in \mathbb{R}^{L \times 2q \times d}$} is current-phase prompt. Similarly, we inject {\small$S$} and {\small ${\bf P}_{T}$} into the decoding procedure of {\small$T$} as follows:
\begin{equation}
\small
H_{T}^l = \text{FFN}\left( \text{Multi-Attn} \left(\textbf{K}_{T}, \textbf{V}_{T}, \textbf{Q}_{T} \right)\right),
\end{equation}\begin{equation}
\small
\textbf{Q}_{T} = H_{T}^{l-1},
\end{equation}
\begin{equation}
\small
\textbf{K}_{T} = [{\bf P}_{T}[l,:q,:]; H_{S}^{l-1}; H_{T}^{l-1}],
\end{equation}
\begin{equation}
\small
\textbf{V}_{T} = [{\bf P}_{T}[l,q:,:];H_{S}^{l-1};H_{T}^{l-1}],
\end{equation}
where {\small$H_{T}^{l}\in \mathbb{R}^{|T|\times d}$} is the decoding state of the $l$-th transformer block. Finally, we refer the $t$-th decoding state as $h_{t}^{L}$ (i.e., {\small $H_{T}^{L}=h_{t}^{L}|_{t=1}^{|T|+1}$}) which is used to predict the next token {$y_t$}: 
\begin{equation}
\label{equ:p-dis}
\small
p\left(y_t|S,\mathcal{C}, y_{<t}\right) = \text{Softmax}\left(h_{t}^{L}W\right),
\end{equation}
where {\small $W\in \mathbb{R}^{d\times |\mathcal{V}|}$} is parameter of LLM-Head layer and {\small $|\mathcal{V}|$} is the vocabulary size.

\subsection{Enhanced Decoding Phase}
\label{sec:cur-dec}
\begin{figure}
\centering
\resizebox{0.7\columnwidth}{!}{
\includegraphics[width=0.7\linewidth, trim=105 0 0 0,clip]{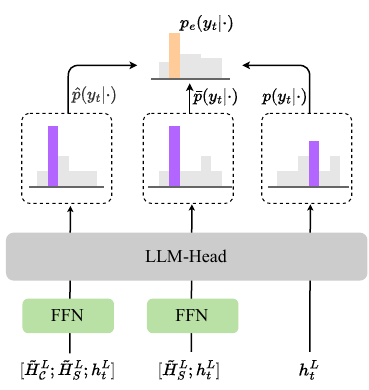}}
\caption{Illustration of the procedure of our proposed decoding-enhanced approach at the $t$-th decoding step.}
    \label{fig:decoding-enhance}
\end{figure}

As shown in Figure~\ref{fig:msp-frame}, both the inter-sentence context representation {\small $H_{\mathcal{C}}^{1:L}$} and the intra-sentence context representation {\small $H_{S}^{1:L}$} are used as keys and values when generating hidden states of next phase. Meanwhile, hidden states of decoding phase, i.e., {\small $h_i^{L}|_{i=1}^{|T|}$} are used to predict next tokens. On the one hand, while the decoding hidden states incorporate both inter- and intra-sentence contexts, there is no explicit differentiation between the two when predicting next tokens. On the other hand, the inter-sentence context representation {\small $H_{\mathcal{C}}^{1:L}$} and decoding hidden states {\small $H_{T}^{1:L}$} are mediated by hidden states of phases 2, i.e., {\small $H_{S}^{1:L}$}. This may result in a \textit{long-distance} issue such that the inter-sentence context are not properly aligned by target-side tokens. 

Therefore, to address above two issues, we propose an enhanced decoding phase with an aim to more effectively utilize both the inter- and intra-sentence contexts. Inspired by~\citet{kuang-etal-2018-attention}, we move both the two types of inter- and intra-sentence contexts closer to target words to achieve a tight interaction between them. Specifically, we concatenate the decoding states with the two types of representations to predict the next target words. As shown in Figure~\ref{fig:decoding-enhance}, the enhanced next word prediction {\small $p_{e}$} is a combination of three distributions:
\begin{equation}
\label{equ:dis_p}
\small
\begin{split}
p_{e}\left(y_t|S,\mathcal{C}, y_{<t}\right) = & \lambda_1\times\hat{p}\left(y_t|S,\mathcal{C}, y_{<t}\right) \\
& + \lambda_2\times\bar{p}\left(y_t|S,\mathcal{C}, y_{<t}\right) \\
& + \left(1-\lambda_1-\lambda_2\right)\times p\left(y_t|S,\mathcal{C}, y_{<t}\right),
\end{split}
\end{equation}
where $\lambda_1$ and $\lambda_2$ control the contribution of $\hat{p}\left(y_t|\cdot\right)$ and $\bar{p}\left(y_t|\cdot\right)$, respectively, which can be further formulated as:
\begin{equation}
\label{equ:p-hat-dis}
\small
\hat{p}\left(y_t|S,\mathcal{C}, y_{<t}\right) = \text{Softmax}\left(\hat{h}_t^{L}W\right),
\end{equation}
\begin{equation}
\label{equ:p-bar-dis}
\small
\bar{p}\left(y_t|S,\mathcal{C}, y_{<t}\right) = \text{Softmax}\left(\bar{h}_t^{L}W\right),
\end{equation}
\begin{equation}
\small
\hat{h}_t^{L} = \text{FFN}\left([\tilde{H}_{\mathcal{C}}^{L}; \tilde{H}_{S}^{L};h_t^{L}]\right),
\end{equation}
\begin{equation}
\small
\bar{h}_t^{L} =\text{FFN}\left( [\tilde{H}_{S}^{L}; h_t^{L}]\right),
\end{equation}
where {\small $W$} is same as in Eq.~\ref{equ:p-dis}, {\small$\tilde{H}_{S}^{L} \in \mathbb{R}^{d}$} and  {\small$\tilde{H}_{\mathcal{C}}^{L} \in \mathbb{R}^{d}$} are the averaged {\small$H_{S}^{L}$} and {\small$H_{\mathcal{C}}^{L}$} at token level, respectively.\footnote{{Notably, the computation of $\hat{p}$ and $\bar{p}$ does not require a full decoding forward pass. It involves solely an FFN layer (two linear transformation layers and a ReLU activation layer), an LLM-Head layer (a linear transformation layer), and a softmax function layer.}} To further identify the effect of inter- and intra-sentence context in this strategy, we provide an ablation study about $\hat{p}$ and $\bar{p}$ in Appendix \ref{apdx:tsl-te-effect}.

\subsection{Phase-aware Prompts}
\label{sec:phase-aware}
We emphasize the LLM needs to play various roles across three phases, and maintaining similar prompts across different phases may not be reasonable. Thus, we empower LLM to distinguish different phases by introducing a type embedding and a transfer layer\footnote{Unlike the multi-layer perceptrons (MLPs) used for reparameterization, our transfer layer shares parameters across all prompts, reducing the number of trainable parameters. Table \ref{tab:params} compares the trainable parameters of various tuning methods, and Appendix \ref{apdx:tsl-te-effect} analyzes the effect of the transfer layer.} for these prompts:
\begin{equation}
\label{equ:trans-embed}
\small
    {\bf P}_{r} = \left(\tanh\left({\bf O}_{r}W_{1}\right)\right)W_{2} + \text{TypeEmb}\left(r\right),
\end{equation}
where {\small${\bf O}_{r}\in \mathbb{R}^{L \times 2q \times d}$} is randomly initialized prompt, {\small$W_1, W_2 \in \mathbb{R}^{d\times d}$} are the trainable parameters, and TypeEmb($\cdot$) is type embeddings layer of the prompts. {\small $r\in \{\mathcal{C}, S, T\}$} represents either phase 1, phase 2, or phase 3.

\subsection{Training Objective}
\label{sec:train-obj}
We employ the cross-entropy loss as the training objective of our model. Given a training instance {\small $(\mathcal{C}, S, T)$}, its training loss is defined as:

\begin{equation}
\small
\mathcal{L}\left(\mathcal{C}, S, T\right) = -\frac{1}{|T|}\sum_{t=1}^{|T|} \text{log}~p_{e}\left(y_t|S,\mathcal{C},y_{<t}\right).
\end{equation}
Notably, the parameters in LLM, including {\small{$W$}} in Eq.~\ref{equ:p-dis}, ~\ref{equ:p-hat-dis}, ~\ref{equ:p-bar-dis}, are frozen during training.
\section{Experimentation}
\begin{table*}
    \centering
    \resizebox{\textwidth}{!}{
    \begin{tabular}{l|rc|rc|rc|rc|rc|rc}
         \toprule
         \multirow{2}{*}{\bf{Model}} & \multicolumn{2}{c|}{\bf ZH$\rightarrow$EN} & \multicolumn{2}{c|}{\bf FR$\rightarrow$EN} & \multicolumn{2}{c|}{\bf DE$\rightarrow$EN} & \multicolumn{2}{c|}{\bf ES$\rightarrow$EN} & \multicolumn{2}{c|}{\bf RU$\rightarrow$EN} & \multicolumn{2}{c}{\bf Average} \\ \cline{2-13}
          & \textit{BLEU} & \textit{COMET} & \textit{BLEU} & \textit{COMET}  & \textit{BLEU} & \textit{COMET}  & \textit{BLEU} & \textit{COMET} & \textit{BLEU} & \textit{COMET} & \textit{BLEU} & \textit{COMET} \\
          \hline
         {$^{\oslash}$}Trans.&29.86  & 0.8406 & 38.53 & 0.8545  & 41.44 & 0.8682 & 48.74 & 0.8783 &32.25  &0.8169 & 38.16 & 0.8517\\
         \hline
         \hline
         \multicolumn{13}{c}{\textit{Traditional context-aware} NMT models } \\
         \hline
         \hline
         {$^{\odot}$}MR-Trans.& 30.61 & 0.8413 & 38.72 &  0.8533 & 42.11 & 0.8693  & 49.69 & 0.8812 & 33.27 &0.8211 & 38.88 &0.8532 \\
         ~ + mBART& 32.69 & 0.8601 & 42.01 &  0.8759 & 44.61 & 0.8840 & 51.67 & 0.8831 & 36.39 & 0.8459& 41.39 &0.8698 \\
        {$^{\odot}$}G-Trans.& 30.99 & 0.8411 &38.96 & 0.8524 & 42.46 &0.8658  & 49.68 &0.8794 & 33.59 & 0.8201 & 39.14 & 0.8518\\
     ~ + mBART& 32.99 & 0.8597 & 42.02 & 0.8764 & 44.81 & 0.8836 & 52.07 & 0.8911 & 36.83 & 0.8461 & 41.74&0.8714 \\
         \hline
         \hline
         \multicolumn{13}{c}{\texttt{llama-2-7b} as foundation model} \\
         \hline
         \hline
         {$^{\oslash}$}MT-LoRA& 27.43 & 0.8511 & 38.18 & 0.8647 & 40.96 & 0.8712 & 47.52 & 0.8733 &33.00  & 0.8311 &37.42& 0.8583\\
         
         {$^{\oslash}$}MT-PT& 31.32 & 0.8565 & 41.92 & 0.8675 & 43.56 & 0.8752 & 51.32 & 0.8819 & 35.46 &0.8333 & 40.72&0.8629\\
         
         {$^{\odot}$}CMT-PT & 31.13 &0.8387  & 42.01 & 0.8699 & 43.11 & 0.8762 & 51.66 & 0.8823 & 35.91 & 0.8396&40.76 &0.8613\\
         \hline
         {$^{\odot}$}{\bf MPT}& *33.21 & 0.8645 & †43.11 & 0.8744 & *43.88 & 0.8824 & †52.01 & 0.8913 & †36.49 &{\bf0.8456} & 41.74& 0.8716\\
         {$^{\odot}$}{\bf DeMPT}& *{\bf33.89} & {\bf 0.8658} & †{\bf43.71} & {\bf0.8816} & *\bf{44.69} & {\bf0.8899} & †\bf{53.10} & {\bf0.8979} & †\bf{36.55} &0.8438 &{\bf 42.39}& {\bf 0.8758}\\
         \hline
         \hline
         \multicolumn{13}{c}{\texttt{bloomz-7b1-mt} as foundation model} \\
         \hline
         \hline
         {$^{\oslash}$}MT-LoRA& 25.79 & 0.8466 & 35.67 & 0.8601 & 35.17 & 0.8522 & 46.32 & 0.8644 & 28.01 &0.8012 &34.21&0.8449 \\
         
         {$^{\oslash}$}MT-PT& 30.99 & 0.8520 & 40.49 & 0.8661 & 37.76 & 0.8579 & 50.68 & 0.8823 & 30.27 &0.8106 &38.04& 0.8539\\
         
         {$^{\odot}$}CMT-PT & 30.82  & 0.8504 & 40.31 &  0.8639&  38.01& 0.8601 & 50.26 & 0.8832 &29.80 &0.8108 & 37.84 & 0.8537\\
         \hline
         {$^{\odot}$}{\bf MPT}&  *31.81& 0.8601 & *41.11 & 0.8766 & †38.99 & 0.8669 & *51.33 & 0.8910 & *30.99 &0.8201&38.85& 0.8629\\
         {$^{\odot}$}{\bf DeMPT}& *\bf{32.46} &  {\bf0.8649}&*{\bf41.92}  & {\bf0.8790} & †{\bf40.06} & {\bf0.8703} & *{\bf52.25} & {\bf0.8990} & *{\bf31.79} & {\bf0.8253} & {\bf 39.70}& {\bf 0.8677}\\
         \bottomrule
    \end{tabular}}
    \caption{Results of different systems on sacreBLEU and COMET metrics. {\bf DeMPT}/{\bf MPT} is our proposed Multi-phase Prompt Tuning approach \textit{with/without} Decoding-enhanced strategy (in Sec. \ref{sec:cur-dec}). Scores with {\bf bold} indicate the best performance. * (or †) indicates the gains are statistically significant over MT-PT (or CMT-PT) with $p$<0.01~\cite{koehn-emnlp-2004-statistical}. {${\oslash}$} and {${\odot}$} indicate the model is \textit{context-agnostic} and \textit{context-aware}, respectively.}
    \label{tab:bleu-comet}
\end{table*}

\begin{table}[h]
    \centering
    \resizebox{\columnwidth}{!}{
    \begin{tabular}{l|c|c|c|c|c|c}
         \toprule
         {\bf{Model}} & {\bf ZH$\rightarrow$} & {\bf FR$\rightarrow$} & {\bf DE$\rightarrow$} & {\bf ES$\rightarrow$} & {\bf RU$\rightarrow$} & {\bf Avg.}\\ 
          \hline
         {$^{\oslash}$}Trans.&47.63  & 54.41 & 58.29 & 62.52  & 48.79 & 54.33\\
         \hline
         \hline
         
          \multicolumn{7}{c}{\textit{Traditional context-aware} NMT model} \\
          \hline
         {$^{\odot}$}MR-Trans.& 48.51 & 55.55 & 59.02 & 63.51  & 49.88 & 55.29\\
         ~ + mBART&50.66 & 58.01 & 61.99 & 66.01 & 54.11 &  58.15\\
         {$^{\odot}$}G-Trans.& 48.99 & 55.31 &  59.23& 63.99  & 50.09 & 55.52\\
         ~ + mBART& 50.98 & 57.88 &  61.97& 66.21 & 54.33 & 58.27 \\
         \hline
         \hline
         \multicolumn{7}{c}{\texttt{llama-2-7b} as foundation model} \\
         \hline
         \hline
         {$^{\oslash}$}MT-LoRA& 44.83 & 54.52 & 57.72 & 62.18 & 49.06 & 53.66\\
         
         {$^{\oslash}$}MT-PT& 49.49 & 57.87 & 60.89 & 65.02 & 52.59 & 57.17\\
         
         {$^{\odot}$}CMT-PT& 49.53 & 58.27 & 61.23 & 65.89 & 53.34 & 57.65\\
         \hline
         {$^{\odot}$}\textit{\bf MPT}& 51.56 & 59.56 & 62.15 & 67.14 & 54.18 & 58.92\\
         {$^{\odot}$}\textit{\bf DeMPT}& {\bf52.68} & {\bf60.33} & {\bf63.11} & {\bf67.95} & {\bf54.94} & {\bf 59.80}\\
         \hline
         \hline
         \multicolumn{7}{c}{\texttt{bloomz-7b1-mt} as foundation model} \\
         \hline
         \hline
         {$^{\oslash}$}MT-LoRA& 43.23 & 51.82 & 51.12 & 61.77 & 43.29 & 50.25\\
         
         {$^{\oslash}$}MT-PT& 49.48 & 56.81 & 55.40 & 64.71 & 46.14& 54.51\\
         
         {$^{\odot}$}CMT-PT& 49.61 & 57.05 & 55.81 & 65.12 & 46.09 & 54.74\\
         \hline
         {$^{\odot}$}{\bf MPT}& 50.22 & 57.93 & 56.69 & 66.25 & 47.29 &55.68\\
         {$^{\odot}$}{\bf DeMPT}& {\bf50.62} & {\bf58.30} & {\bf57.34} & {\bf67.12} & {\bf48.00} & {\bf 56.28}\\
         \bottomrule
    \end{tabular}}
    \caption{Results of different systems on BlonDe metric.}
    \label{tab:BlonDe}
\end{table}

We build our approach upon two open-source LLMs, i.e., \texttt{llama-2-7b}\footnote{\url{https://huggingface.co/meta-llama/Llama-2-7b-hf}} and \texttt{bloomz-7b1-mt}\footnote{\url{https://huggingface.co/bigscience/bloomz-7b1-mt}}. We verify the effectiveness of our proposed approach on five translation tasks, including \{Chinese (ZH), French (FR), German (DE), Spanish (ES), Russian (RU)\}$\rightarrow${English (EN)}.

\subsection{Experimental Settings}
\paragraph{Datasets and Preprocessing.} The corpus of all translation tasks is extracted from {\texttt{News-Commentary-v18}}. For LLM-based models, We use the tokenizer of foundation models to process the input data and no other preprocessing is performed. See Appendix \ref{apdx:data} for more details on splitting, preprocessing and statistics of datasets. {Besides, we provide a discussion for scales of the training set in Appendix \ref{apdx:dataset-scale} .}

\paragraph{Baselines.} In addition to \textit{traditional context-agnostic} (Trans.) and \textit{context-aware} (G-Trans~\cite{bao_etal_acl_2021_gtrans} and MR-Trans~\cite{sun_etal_2022_acl_rethinking} with or without pre-training setting, i.e., + mBART~\cite{liu2020multilingual}) NMT models {with \textit{encoder-decoder} architecture} ,\footnote{Please refer to Appendix \ref{apdx:i-gtrans-mr} for more introduction about the G-Trans and MR-Trans.} our primary comparison focuses on the following three LLM-based alternatives: 1) \textbf{MT-LoRA}: It is a tuned LLM adapted to NMT task via the tuning method of Low-Rank Adaptation \cite{hu2022lora}, which makes large-scale pre-training models adapt to a new task by injecting a trainable rank decomposition matrice into each layer of the Transformer architecture; 2) \textbf{MT-PT}: It is a tuned LLM adapted to NMT task via the deep prompt tuning with MLPs reparameterization,\footnote{We attempt to remove reparameterization but experience a significant decline in performance.} which only tunes continuous prompts with a frozen language model; 3) {\bf CMT-PT}: It indiscriminately utilizes inter- and intra-sentence context via the concatenation strategy, as depicted in Figure~\ref{fig:concat-phase} (a). Similar to MT-PT, it is also a tuned LLM via the deep prompt tuning with MLPs reparameterization. Among them,  MT-LoRA and MT-PT are \textit{context-agnostic} systems while CMT-PT is a \textit{context-aware} system. For a fair comparison, we ensure that all context-aware models built upon LLM, including CMT-PT, MPT, and DeMPT, incorporate identical inter-sentence context. We provide more discussion in utilization of various inter-sentence contexts in Appendix \ref{apdx:ctx-effect} and \ref{apdx:tgt-ctx}.

\paragraph{Model Setting and Training.} For all \textit{encoder-decoder} Transformer models, including Transformer (Trans.), MR-Trans and G-Trans\footnote{For G-Trans, we use their official implementation upon Fairseq. Code: \url{https://github.com/baoguangsheng/g-transformer}.}, we implement them upon Fairseq \cite{ott2019fairseq}. For MT-LoRA models, we set the rank of trainable matrices as 16 which performs best in our preliminary experiment. For all MT-PT models, CMT-PT models, and our models, we set the prompt length $q$ as 64.\footnote{We provide more discussion in Appendix \ref{apdx:p-len-ef} about the prompt length.} For the incorporation of inter-sentence context in CMT-PT and our models, we consider a dynamic $z$, in which the total tokens are no more than 256. In enhanced decoding, we consider the three next word predictions to be equally important by setting both {\small $\lambda_1$} and {\small $\lambda_2$} to 1/3. {We provide an analysis of $\lambda$  and more training details in Appendix \ref{apdx:lambda} and \ref{apdx:training}, respectively.}

\paragraph{Evaluation.} We use sacreBLEU (accuracy-related metric)\footnote{Signature: {\tt nrefs:1|case:mixed|eff:no|tok:13a|\\smooth:exp|version:2.3.1}} \cite{post-2018-wmt-call}, COMET (semantics-related metric) with the \texttt{wmt22-comet-da} model\footnote{\url{https://github.com/Unbabel/COMET}} ~\cite{rei-etal-2020-comet}, and BlonDe (discourse-related metric) \cite{jiang-etal-2022-BlonDe} as evaluation metrics.\footnote{We provide more discourse-related evaluation in Appendix \ref{apdx:contras-set-perf}.}

\subsection{Experimental Results}
The main experimental results are presented in Tables \ref{tab:bleu-comet} and \ref{tab:BlonDe}. Additionally, a comparison of the number of trainable parameters is presented in Table \ref{tab:params} across different tuning methods. When examining \texttt{llama-2-7b} and focusing on context-agnostic models, we find that the Transformer models (Trans.) generally outperform LLMs with LoRA tuning (MT-LoRA) in most translation directions based on BLEU score. However, the MT-LoRA models surpass Trans. in COMET, indicating that translations from LLMs may better align with human preferences. Additionally, the MT-PT models exhibit superior performance compared to the MT-LoRA models across BLEU, COMET, and BlonDe metrics. This improvement could be attributed to the more trainable parameters in the MT-PT models (13.87\% vs. 0.12\%).

Importantly, by comparing MT-PT and CMT-PT, we observe that CMT-PT which indiscriminately leverages the inter- and intra-sentence context with the concatenation way, even hurts performance for certain translation tasks. For example, the CMT-PT models, despite excelling in discourse-related BlonDe scores (averaging 57.65 vs. 57.17), underperforms in BLEU and COMET compared to the MT-PT models. In contrast, our context-aware MPT and DeMPT models outperform all LLM baselines across all translation tasks in three metrics. For example, our MPT models achieve an average gain of 0.98/0.0103/1.27 in BLEU/COMET/BlonDe compared to the CMT-PT models. Our decoding-enhance strategy further enhances the capacity of LLMs, with DeMPT outperforming MPT with an average gain of 0.65/0.0042/0.88. Compared to G-Trans. (+mBART) or MR-Trans (+mBART), DeMPT also demonstrates either superior or comparable performance across all language pairs.

Finally, we observe a similar performance trend among MT models built upon \texttt{bloomz-7b1-mt}. It also indicates that models built upon \texttt{llama-2-7b} outperform those utilizing \texttt{bloomz-7b1-mt}, suggesting that \texttt{llama-2-7b} serves as a more robust foundation model for translation tasks.

\begin{table}
    \centering
    \resizebox{\columnwidth}{!}{
    \begin{tabular}{l|c|c|c}
         \toprule
         {} & {\bf MT-LoRA} & {\bf MT-PT/CMT-PT} & {\bf MPT/DeMPT}  \\
          
         \hline
         Trainable Para. &0.12\%  & 13.87\% & 3.11\%  \\
         \bottomrule
    \end{tabular}}
    \caption{Proportion of trainable parameters against total parameters for different tuning methods.}
    \label{tab:params}
\end{table}
\section{Discussion}
In this section, we use \texttt{bloomz-7b1-mt} as the foundation model to discuss our approach.\footnote{{Considering page limitation and the consumption of GPUs resources and training time, we use the ZH$\rightarrow$EN task as a representative to report the BLEU and BlonDe scores.}} See Appendix \ref{apdx:speed}$\sim$\ref{apdx:tgt-ctx} for further discussions.
\subsection{Effect of Length of Inter-sentence Context}

\begin{figure}
\centering
\resizebox{\columnwidth}{!}{
\includegraphics[width=1.0\linewidth, trim=5 54 14 85,clip]{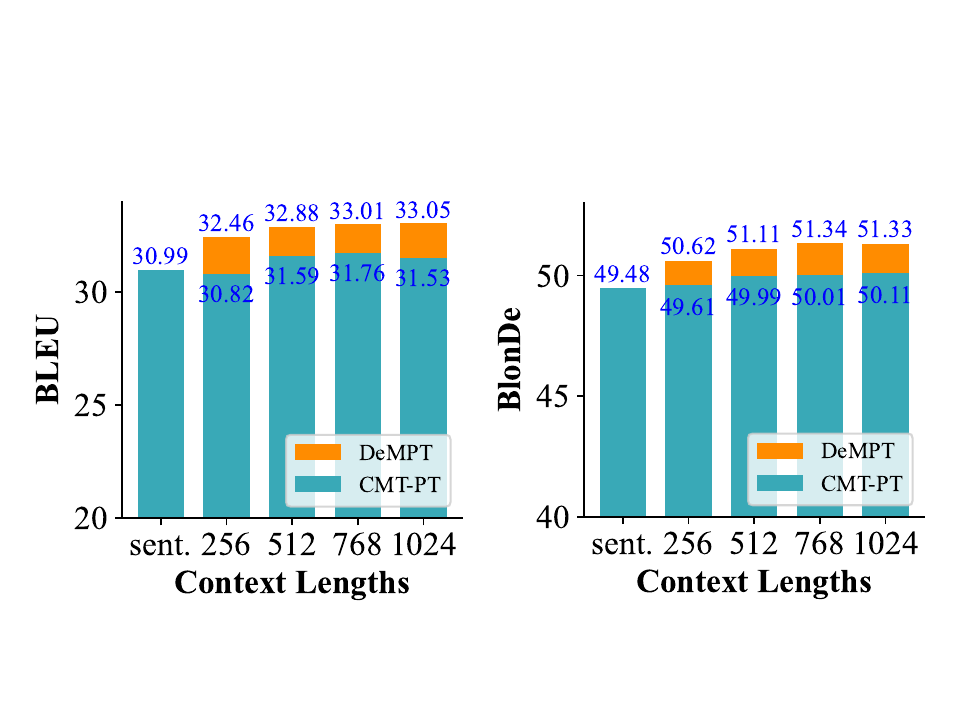}}
    \caption{Performance of CMT-PT and our DeMPT on ZH$\rightarrow$EN test set when using different inter-sentence context lengths.}
    \label{fig:our-lengths-effect}
\end{figure}

For efficient training, we define the inter-sentence context in Section~\ref{sec:appro} as previous sentences with a total tokens not exceeding 256. We are curious about the potential impact of inter-sentence length on the performance of our approach. Consequently, we extend the inter-sentence context length from 256 to 1024 and assess the performance of our approach in the ZH$\rightarrow$EN task. Figure \ref{fig:our-lengths-effect} shows the performance trend of the CMT-PT model and our DeMPT model. As the length of the inter-sentence context increases, both models exhibit a slight enhancement in both BLEU and BlonDe scores. Interestingly, our model with a 256-token inter-sentence context outperforms the CMT-PT model with a 1024-token inter-sentence context in both BLEU and BlonDe scores. This further suggests the effectiveness of our approach in harnessing the capabilities of LLMs for context-aware NMT compared to the concatenation strategy.

\subsection{Effect of Multi-phase Strategy}
Our multi-phase strategy divides the whole translation into three phases: phase 1 for encoding inter-sentence, phase 2 for encoding intra-sentence, and phase 3 for decoding current sentence. To assess the effect of multi-phase strategy, we compare its performance with two contrasting strategies: merging the first two phases into one (i.e., \textit{Merging 1\&2}) and and merging all phases into a single one (i.e., \textit{Merging 1\&2\&3}).\footnote{{When merging them all into one, it equals CMT-PT, i.e., the concatenate strategy.}} Note that in both the two contrasting strategies, we replace the enhanced next word prediction $p_e\left(y_t|\cdot\right)$ (Eq.~\ref{equ:dis_p}) in decoding phase with $p\left(y_t|\cdot\right)$. Table~\ref{tab:merged-phases} presents the performance of different phrasing strategies. Comparing \textit{Merging 1\&2} and \textit{Merging 1\&2\&3}, it shows that separating the decoding phrase from the encoding marginally improves the performance in both BLEU and BlonDe. Importantly, the comparison of MPT and \textit{Merging 1\&2} tells that separating the encoding of inter- and intra sentence achieves more gains across all metrics. 
\begin{table}[]
\resizebox{\columnwidth}{!}{
\begin{tabular}{lccc}
\toprule
{\bf Model } &{\bf BLEU} &{\bf COMET} &{\bf BlonDe} \\
\hline
MPT  &31.81& 0.8601 &50.22 \\
DeMPT &32.46 &0.8649 &50.62  \\ 
\hdashline
\textit{Merging 1\&2\&3} & 30.82 &0.8504 & 49.61  \\
\textit{Merging 1\&2} &31.01 & 0.8503 &49.91\\
\bottomrule
\end{tabular}}
\caption{Comparison of performances when using different phrasing strategies on ZH$\rightarrow$EN test set.}
\label{tab:merged-phases}
\end{table}

Meanwhile, we conjecture that another benefit of multi-phasing strategy lies in the robustness to the noise contained in document-level context. To test the conjecture, we replace the original inter-sentence context with a random inter-sentence context, meaning we randomly select some sentences from other documents to serve as the inter-sentence context. As shown in Table~\ref{tab:random-ctx}, the performance of both the \textit{Merging 1\&2\&3} and DeMPT models consistently deteriorates when exposed to random context (\textit{w/} rnd. CTX). However, the decline is more pronounced for \textit{Merging 1\&2\&3} than for DeMPT (-2.19/0.0102/1.60 vs -0.90/0.0068/0.91). This suggests that DeMPT, owing to its multi-phase strategy, exhibits more robustness in utilizing inter-sentence context in contrast to \textit{Merging 1\&2\&3}. 

\begin{table}[]
\resizebox{\columnwidth}{!}{
\begin{tabular}{lccc}
\toprule
{\bf Model } &{\bf BLEU} &{\bf COMET} &{\bf BlonDe} \\
\hline
\textit{Merging 1\&2\&3} &   30.82 &0.8504 &49.61  \\
~ \textit{w/} rnd. CTX & 28.63 & 0.8402 & 48.01   \\
\hdashline
DeMPT &32.46 &0.8649 &50.62  \\ 
~ \textit{w/} rnd. CTX  &31.56 &0.8581 &49.71 \\
\bottomrule
\end{tabular}}
\caption{Comparison of performance when using gold or random inter-sentence context on ZH$\rightarrow$EN test set.}
\label{tab:random-ctx}
\end{table}

\subsection{Human Evaluation}
\label{sec:da}
We use the Direct Assessment (DA) method \cite{graham-etal-2017-da} to manually assess the quality of translations generated by DeMPT and CMT-PT. In this assessment, human evaluators compare the meaning of the MT output with a human-produced reference translation, working within the same language. Specifically, we randomly select 5 documents with a total of 200 groups of sentences from the ZH$\rightarrow$EN test set. To avoid potential bias in evaluation, we recruit 6 professional translators and ensure each translation from DeMPT or CMT-PT is scored twice by two translators. Table \ref{tab:da-score} shows the DA scores for CMT-PT and DeMPT. Our DeMPT outperforms CMT-PT by 7.14 DA score, providing strong evidence for the effectiveness of our approach. Further details and results regarding the DA can be found in {Appendix} \ref{apdx:da}.
\begin{table}[t]
    \centering
    \resizebox{\columnwidth}{!}{
    \begin{tabular}{llll}
         \toprule
         {\bf{Model}} & {\bf Score\_1} & {\bf Score\_2} & {\bf Average} \\
         \hline
         CMT-PT &79.00 & 80.17 & 79.59\\
         DeMPT & \bf{86.17} {\small(\textcolor{blue}{+7.17})}& \bf{87.30} {\small(\textcolor{blue}{+7.13})} &\bf{86.73 {\small(\textcolor{blue}{+7.14})}}\\
         \bottomrule
    \end{tabular}}
    \caption{Human DA scores for CMT-PT and DeMPT on ZH$\rightarrow$EN translation task.}
    \label{tab:da-score}
\end{table}
\section{Related Work}
Due to limited space, we omit the discussion on conventional context-aware MT, focusing instead on LLM-based context-aware MT and prompt tuning for LLMs. {Besides, considering our DeMPT's inspiration from MSP~\cite{tan-etal-2022-msp}, we offer further discussion on their differences.}

\paragraph{LLM-based Context-aware Machine Translation.} While traditional context-aware neural machine translation (NMT) has seen considerable progress in recent years \cite{jean_etal_2017,wang_etal_emnlp_2017,voita_etal_acl_2018,maruf_etal_naacl_2019,kang_etal_emnlp_2020,bao_etal_acl_2021_gtrans,sun_etal_2022_acl_rethinking,bao-etal-2023-targetda}, the effective integration of large language models (LLMs) to model inter-sentence context and enhance context-aware translation remains an area of limited exploration. Existing studies mainly focus on the assessment of LLMs' ability in discourse modeling. For example, \citet{wang-etal-2023-document-level} approach context-aware NMT as a task involving long sequence generation, employing a concatenation strategy, and conduct comprehensive evaluations of LLMs such as ChatGPT and GPT-4. Their focus includes the impact of context-aware prompts, comparisons with translation models, and an in-depth analysis of discourse modeling ability. Similarly, \citet{karpinska-iyyer-2023-large} engage professional translators to evaluate LLMs' capacity in context-aware NMT.  In contrast, \citet{Wu2024AdaptingLL} compare the effectiveness of various parameter-efficient fine-tuning methods on moderately-sized LLMs for context-aware NMT. Besides, \citet{wu-hu-2023-exploring} explore the prompt engineering with GPT language models specifically for document-level (context-aware) MT while \citet{Li2024Enhancing} experiment with combining sentence-level and document-level translation instructions of varying lengths to fine-tune LLMs. {Differently, \citet{koneru_etal_2024_contextual} propose a post-editing approach to enhance LLMs' capacity in utilization of inter-sentence context in document-level MT.}

\paragraph{Prompt Tuning for Large Language Model.} \citet{liu2021gptudt} and \citet{li-liang-2021-prefix} propose to make LLMs adapt to various tasks by adding trainable prompts (also called continuous prompts) to the original input sequences. In this paradigm, only the continuous prompts are updated during training. \citet{liu2022pt-v2} further introduces deep prompt tuning, extending the idea by inserting trainable prompts into all layers of LLMs, rather than just the embedding layer. While these approaches provide a general framework, we focus on enhancing LLM performance specifically for inter-sentence context modeling in context-aware NMT.

\paragraph{Discussion with MSP.} \citet{tan-etal-2022-msp} propose a multi-phase tuning approach (MSP) to enhance the sentence-level translation performance of a multilingual GPT. Our DeMPT mainly differs from MSP in the following aspects:
 1) DeMPT adopts a phase-aware prompt to enable distinctive modeling for different inputs, namely inter-sentence contexts, intra-sentence contexts, and the target sentence, a feature not present in MSP;
2) DeMPT incorporates a decoding-enhanced strategy to further improve the effectiveness of utilizing different context information, a capability not available in MSP;
3) DeMPT is designed to alleviate discourse problems in context-aware LLM-based machine translation tasks, rather than addressing sentence-level machine translation tasks as in the case of MSP;
4) DeMPT is designed to adapt LLMs rather than smaller pre-trained model used in MSP.

\section{Conclusion}
In this paper, we have examined the hypothesis that it is crucial to differentially model and leverage inter-sentence context and intra-sentence context when adapting LLMs to context-aware NMT. This stems from our observation that intra-sentence context exhibits a stronger correlation with the target sentence compared to inter-sentence context, owing to its richer parallel semantic information.
To this end, we have proposed a novel decoding-enhanced multi-phase prompt tuning (DeMPT) approach to make LLMs aware of the differences between inter- and intra-sentence contexts, and further improve LLMs' capacity in discourse modeling. We have evaluated our approach using two foundation models and present experimental results across five translation directions. Experimental results and discussions have demonstrated a significant enhancement in the performance of LLMs in context-aware NMT, manifesting as improved translation accuracy and a reduction in discourse-related issues.
\section*{Limitations}
Owing to resource limitations, our work is restricted to moderate-scale LLMs, specifically those with 7 billion parameters, and a confined window size of inter-sentence context. It is imperative to acknowledge that the results of our research may differ when employing larger models and extended window sizes for inter-sentence contexts. Considering that English text forms the main body of the training data for LLMs, this paper only focuses on the English-centric translation tasks. The results of non-English-centric translation tasks may vary. We acknowledge these limitations and consider them as avenues for future exploration. {Besides, following the finding of \citet{bao_etal_acl_2021_gtrans}, we focus solely on the source-side inter-sentence context in this work. We will explore more about the integration of target-side inter-sentence context in the future.}

\bibliography{custom}

\appendix
\label{sec:appendix}
\begin{CJK*}{UTF8}{gkai}
\section{Datasets}
\label{apdx:data}
\paragraph{Splitting, Preprocessing and Statistics of Datasets.} For all translation tasks, we randomly select 80\% document pairs from the corpus as the training set. Both the test set and validation set include 150 document pairs each, randomly sampled from the remaining 20\% of document pairs in the corpus. Regarding sentence preprocessing across all datasets for LLM-based models, we segment the sentences with the tokenizer from the respective foundation model. No additional preprocessing steps are performed. For \textit{encoder-decoder} Transformer models, we segment the source and target sentences into sub-words by a
BPE model with 30K merged operations \cite{sennrich_etal_acl_2016}. We provide the detailed statistic in Table \ref{tab:datastst}. Datasets are downloaded from {\url{https://data.statmt.org/news-commentary/v18}.}

\begin{table*}[]
\centering
\resizebox{\textwidth}{!}{
\begin{tabular}{l|rr|rr|rr|rr|rr}
\toprule
\multirow{2}{*}{\bf Dataset} & \multicolumn{2}{c|}{\bf ZH$\rightarrow$EN} & \multicolumn{2}{c|}{\bf FR$\rightarrow$EN} & \multicolumn{2}{c|}{\bf DE$\rightarrow$EN} & \multicolumn{2}{c|}{\bf ES$\rightarrow$EN} & \multicolumn{2}{c}{\bf RU$\rightarrow$EN}\\
&\#Doc &\#Sent &\#Doc &\#Sent&\#Doc &\#Sent&\#Doc &\#Sent&\#Doc &\#Sent \\
\cline{1-11}
Training&8,622 &342,495 &7,915 &310,489 &8,417 &333,201 &9,677 &378,281 &7,255 &272,100 \\
Validation&150 &6,061 &150 &5,890 &150 &5,866 &150 & 5,782 &150 & 5,691 \\
Test &150 &5,747 &150 &5,795 &150 &5,967 &150 &5,819 &150 & 5,619 \\  
\bottomrule
\end{tabular}}
\caption{Statistics of training, validation, and test sets for five translation tasks. \#Doc and \#Sent denote the numbers of \textit{Document} and \textit{Sentence}, respectively.}
\label{tab:datastst}
\end{table*}

\begin{figure*}[h]
    \centering
    \resizebox{0.9\textwidth}{!}{
\includegraphics{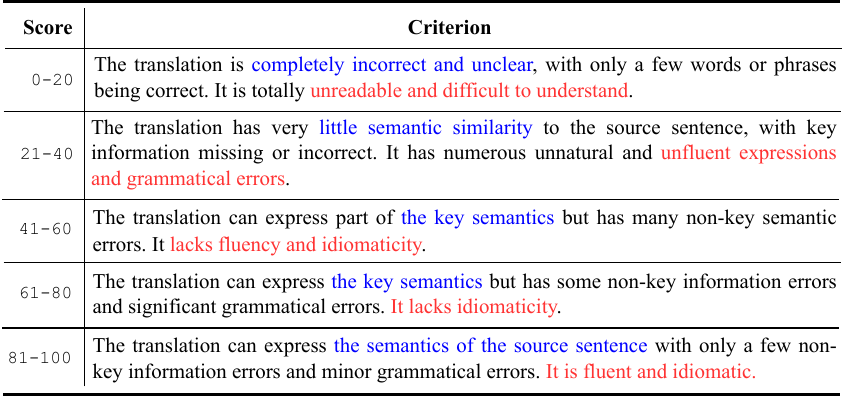}}
    \caption{Scoring criterion for Direct Assessment. We group the score into five ranges, i.e., \texttt{0-20}, \texttt{21-40},  \texttt{41-60},  \texttt{61-80}, \texttt{81-100}.}
    \label{fig:da-criterion}
\end{figure*}
\section{Training Details}
\label{apdx:training}
For all \textit{encoder-decoder} Transformer NMT models, we use the transformer-base setting as in \citet{vaswani_etal_nips_2017}, where the learning rate is set to $1e-4$ with an inverse-square schedule and warmup steps of $4000$, and use Adam optimizer with $\beta_{1}=0.9$ and $\beta_{2}=0.98$. For the other special training settings in G-Trans and MR-Trans, we keep consistent with that provided in their paper. All Transformer NMT models are trained on 4$\times$ NVIDIA V100 32GB GPUs with a batch size of 4096. For the models with prompt tuning in Section \ref{sec:exp}, including MT-PT, CMT-PT, MPT and DeMPT models, the length of the trainable prompt is set as 64. During both training and inference, the model generates only the current target sentence, operating in a many-to-one translation mode. For all fine-tuning models in this paper, we set the training epoch to 4, and the warm-up rate to 0.1. We use the log learning rate decay strategy with a maximum learning rate of 5e-5. We collate a mini-batch by counting the total tokens inside the batch and set the batch size as 4096. All fine-tuning models are trained on 4 $\times$ NVIDIA A800 GPUs with Deespeed Zero 2 offload setting \cite{Rajbhandari-etal-2020-deepspeed}.\footnote{\url{https://github.com/microsoft/DeepSpeed}}
\begin{figure*}[h]
    \centering
    \resizebox{\textwidth}{!}{
    \includegraphics{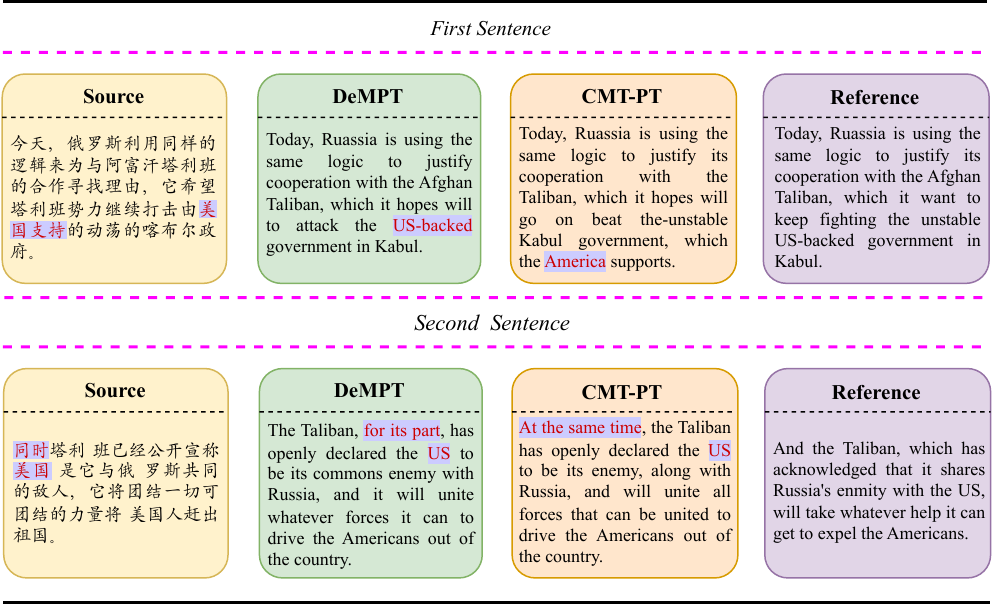}}
    \caption{A case study for the CMT-PT model and our DeMPT model on ZH$\rightarrow$EN translation task.}
    \label{fig:case-study}
\end{figure*}

\section{Traditional Context-aware Models}
\label{apdx:i-gtrans-mr}
In this paper, we implement G-Transformer (G-Trans) \cite{bao_etal_acl_2021_gtrans} and Multi-Resolution Transformer (MR-Trans) \cite{sun_etal_2022_acl_rethinking} as representatives of traditional context-aware models for comparison. For ease of understanding, we provide a brief introduction to these two models in this section.

\paragraph{G-Transformer.} The transformer model equips a group attention on the lower encoder/decoder layer and a combined attention on the top encoder/decoder layer. For a sentence being translated with its inter-sentence context, the group attention helps maintain locality bias by focusing on intra-sentence context. Meanwhile, the combined attention effectively integrates boundary information, enhancing the translation process with inter-sentence context.

\paragraph{Multi-Resolution Transformer.} The Transformer model does not include any additional modules specifically for modeling inter-sentence context. Instead, it only uses a mixed training set that comprises both sentence-level and document-level instances with varying numbers of sentences. Training on this mixed set allows the Transformer model to handle both sentence-level and document-level translation tasks. In this paper, we implement its Document-to-Sentence variant, which uses all preceding contexts as the source and the current sentence as the target.

\section{Comparison of Inference Speed}
\label{apdx:speed}
\begin{table}[ht]
\centering
\begin{tabular}{lll}
\toprule
\bf{Model}  & {\bf Speed} & {\bf BLEU} \\
\hline
MT-PT  & 0.75 \textit{sec/sent.}  & 30.99    \\
CMT-PT & 0.77 \textit{sec/sent.}  & 30.82   \\
MPT   & 0.78 \textit{sec/sent.}  & 31.81     \\
DeMPT  & 0.79 \textit{sec/sent.}  & 32.46   \\
\bottomrule
\end{tabular}
\caption{Comparison of inference speed on ZH$\rightarrow$EN translation task. Speed is measured on the test set using 4 GPUs. \textit{sec/sent.} means seconds spent for decoding each sentence. Note that the reparameterization is not needed during inference \cite{li-liang-2021-prefix}.}
\label{tab:inf-speed}
\end{table}
Table~\ref{tab:inf-speed} compares the inference speed of different models on ZH$\rightarrow$EN translation task. Our MPT and DeMPT models, dividing the context-aware NMT process into three separate phases, demonstrates comparable inference speed to the single-phase MT-PT and CMT-PT models, with only a marginal drop of 0.02 seconds per sentence in decoding. This illustrates the efficiency of our approach without introducing significant computational overhead. 

\section{Effect of Prompt Length}
\label{apdx:p-len-ef}
\begin{figure}[t]
\centering
\resizebox{\columnwidth}{!}{
\includegraphics[width=1.0\linewidth, trim=18 51 7 85,clip]{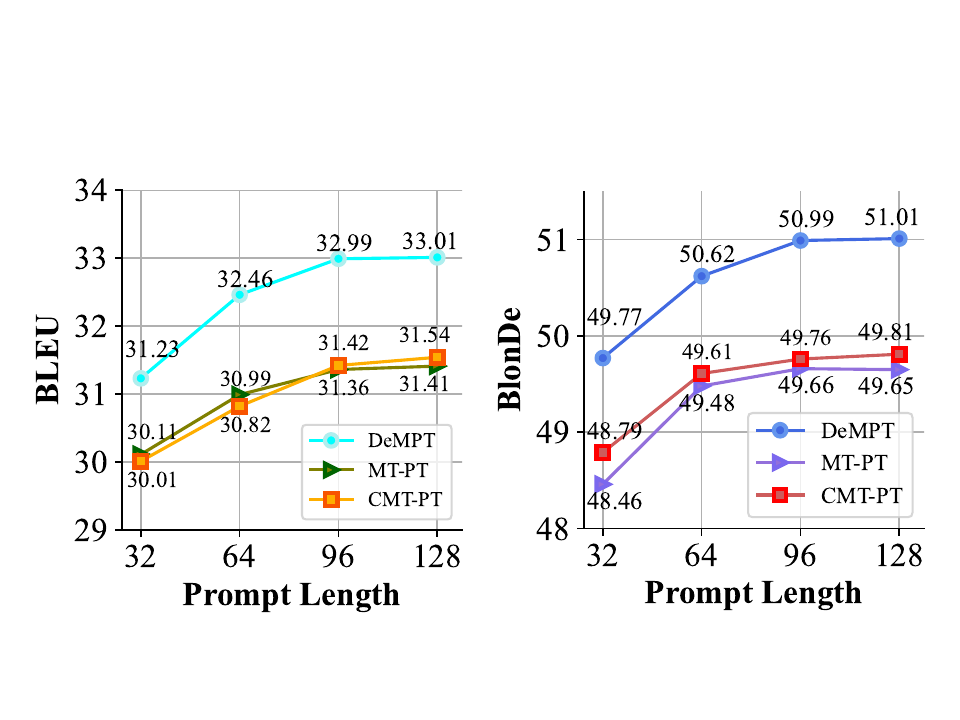}}
       \caption{Performance of MT-PT, CMT-PT, and our DeMPT on ZH$\rightarrow$EN test set when using different lengths of the trainable prompts.}
    \label{fig:pt-len}
\end{figure}
As our approach is implemented based on deep prompt tuning, next we compare the impact of the trainable prompt length for MT-PT, CMT-PT, and our DeMPT. Figure \ref{fig:pt-len} shows the performance curves when increasing the prompt length from 32 to 128. We observe that increased prompt length tends to enhance performance for both BLEU and BlonDe, yet the gains exhibit diminishing returns. This finding is consistent with that in \citet{li-liang-2021-prefix,lester-etal-2021-power,tan-etal-2022-msp}. We also observe that DeMPT with a prompt length of 64 outperforms both MT-PT and CMT-PT with a prompt length of 128 on both metrics, suggesting the superiority of our approach over the concatenation strategy in enhancing LLMs' capacity for context-aware NMT.

\section{Performance on Contrastive Test Set}
\label{apdx:contras-set-perf}
We evaluate the models' ability to resolve discourse inconsistencies using the contrastive test set proposed by~\cite{voita_etal_emnlp_2019}, which focuses on four discourse phenomena such as deixis, lexicon consistency (lex.c), ellipsis inflection (ell.infl), and verb phrase ellipsis (ell.VP) in English$\rightarrow$Russian translation. Within the test set, each instance comprises a positive translation and several negative ones that vary by only one specific word. The purpose of the contrastive test set is to assess whether a model is more inclined to generate a correct translation as opposed to incorrect variations. Table \ref{tab:contrative-set} lists the accuracy of translation prediction on the contrastive test set for MT-PT, CMT-PT and DeMPT. Compared to the context-agnostic MT-PT model, both context-aware CMT-PT and DeMPT models show substantial improvements across the four discourse phenomena. Additionally, DeMPT demonstrates the best performance, surpassing CMT-PT by an average accuracy margin of 3.8.

\begin{table}[t]
    \centering
    \resizebox{\columnwidth}{!}{
    \begin{tabular}{lrrrrr}
         \toprule
         {\bf{Model}} & {\bf deixis} & {\bf lex.c} & {\bf ell.infl} & {\bf ell.VP} & {\bf Avg.}\\
         \hline
         MT-PT &50.0  & 45.7 & 53.0 & 28.6 & 44.3\\
         CMT-PT &\bf{80.2}  & 46.1 & 74.3 & 75.3 & 68.9\\
         \hline
         DeMPT&80.1  & \bf{55.7} & \bf{75.9} & \bf{79.3} &\bf{72.7}\\
         \bottomrule
    \end{tabular}}
    \caption{Accuracy [\%] of translation prediction for four discourse phenomena on the English $\rightarrow$ Russian contrastive test set.}
    \label{tab:contrative-set}
\end{table}

\section{Effect of Various Contexts for Decoding-enhanced Strategy}
\label{apdx:peffect}

\begin{table}[h]
\centering
\begin{tabular}{lccc}
\toprule
{\bf Model} &{\bf BLEU} &{\bf COMET} &{\bf BlonDe} \\
 \hline
MT-PT  &30.99 &0.8520 &49.48 \\
CMT-PT &30.82 &0.8504 &49.61  \\
\hdashline
DeMPT  &32.46 &0.8649 & 50.62 \\
 ~~~\textit{w/o} $\hat{p}$& 32.33 & 0.8629 & 50.29 \\
 ~~~\textit{w/o} $\bar{p}$& 32.11 & 0.8641 & 50.51 \\
\bottomrule
\end{tabular}
\caption{Comparison of performances of the DeMPT when removing different probabilities $p$ in decoding-enhanced strategy.}
\label{tab:ablation-p}
\end{table}

We conduct an ablation study on the ZH-EN translation direction using the \texttt{bloomz-7b-mt} model as the foundation model to clarify the effect of the three probabilities $p$ in Equation \ref{equ:dis_p}, i.e., the effect of various contexts for the heuristic decoding-enhanced strategy. From the Table \ref{tab:ablation-p}, we observe that removing $\hat{p}$, i.e., \textit{w/o} $\hat{p}$, leads to a significant degradation in the discourse-related metric, namely the BlonDe. This is because the integration enhances the utilization of the inter-sentence context during the decoding phase. We are additionally, removing results in the most substantial degeneration in BLEU metric. This observation demonstrates that our heuristic decoding-enhanced strategy can distinctively improve the utilization of various contexts during the decoding phase.

\section{Details of Human Evaluation}
\label{apdx:da}
\paragraph{Criterion and Recruitment.} Given a source sentence, its translation from MT (i.e., CMT-PT and our DeMPT), and its human-produced reference translation, the evaluators are asked to give a score ranging from 0 to 100. Figure \ref{fig:da-criterion} presents the detailed criterion of scoring. We recruit evaluators from professional translators with at least five years of experience in translation.

\paragraph{Statistics of Translation Errors.} We manually count the number of bad cases from our DeMPT model. The bad cases fall into two categories:  (1) the DA score is 60 or lower; (2) the DA score is lower than that of the translation from CMT-PT. The main types of the bad cases are \textbf{Mistranslation } (\texttt{Mis.}), \textbf{Unnoticed Omission} (\texttt{UO}), \textbf{Inappropriate Expression} (\texttt{IE}), and \textbf{Grammatical Error} (\texttt{GE}). We present detailed statistics in Table \ref{tab:error-stst}. The statistics indicate the bad cases mainly come from Mistranslation and Unnoticed Omission. Meanwhile, our DeMPT model outperforms the CMT-PT model in 86.5\% DA cases.
\paragraph{Case Study.} We present a case in {Figure \ref{fig:case-study}} to illustrate how our DeMPT model outperforms the CMT-PT model. In this case, we compare the translations of two consecutive sentences from our model and the CMT-PT model. First, we notice that the CMT-PT model translates the source word 美国 ~in the two sentences into \textit{US} and \textit{America}, respectively. However, our model \textbf{consistently} translates them into \textit{US}. Second, our model uses \textit{for its part}, a phase with more \textbf{coherent preference}, as the translation of 同时~, instead of \textit{At the same time} adopted in the translation from the CMT-PT model. Both of them demonstrate the superiority of our proposed approach in discourse modeling.
\begin{table}[t]
\resizebox{\columnwidth}{!}{
\begin{tabular}{c|ccccl}
\toprule
\multicolumn{1}{c|}{\multirow{2}{*}{\bf Group}} & \multicolumn{5}{c}{ \bf Type of Bad Case}                                                         \\ \cline{2-6} 
\multicolumn{1}{c|}{}                       & \multicolumn{1}{l}{\texttt{Mis.}} & \texttt{UO} & \texttt{IE} & \texttt{GE} & \textit{Total (Perc.)} \\ \cline{1-6}
1 &6 &3 &1 &2 &12 (\textbf{6.0\%}) \\ 
2 &9 &7 &6 &5 &27 (\textbf{13.5\%})              \\ 
\bottomrule
\end{tabular}}
\caption{Statistics of bad cases from our DeMPT model on ZH$\rightarrow$EN translation task. \textit{Perc.} denotes the percentage of bad cases against the total of DA cases.}
\label{tab:error-stst}
\end{table}

\section{Effect of Dataset Scales}
\label{apdx:dataset-scale}
\begin{table}[]
\resizebox{\columnwidth}{!}{
\begin{tabular}{lccc}
\toprule
{\bf Model } &{\bf BLEU} &{\bf COMET} &{\bf BlonDe} \\
\hline
CMT-PT & 30.82 &0.8504 &49.61  \\
~ + 200K  & 31.21& 0.8521 & 49.88   \\
~ + 400K  & 31.73 & 0.8555 &50.11\\
~ + 700K  &31.89	&0.8559	&50.23
\\
\hdashline
DeMPT &32.46 &0.8649 &50.62  \\ 
~ + 200K  & 32.77 & 0.8663 & 50.99   \\
~ + 400K  & 33.56& 0.8701 & 51.47\\
~ + 700K  &33.91	&0.8721	&51.97\\
\bottomrule
\end{tabular}}
\caption{Comparison of performances of CMT-PT and DeMPT trained on the different scales of corpus for the ZH$\rightarrow$EN translation task.}
\label{tab:datascales}
\end{table}

We conduct an experiment to analyze the impact of training dataset scales on the concatenating strategy (CMT-PT) and the multi-phased, decoding-enhanced strategy (DeMPT). To do this, we expand the ZH$\rightarrow$EN training set with additional document-level data from the LDC.\footnote{The training data set consists of LDC2002T01, LDC2004T07,
LDC2005T06, LDC2005T10, LDC2009T02,
LDC2009T15, and LDC2010T03.} Specifically, we selected 200K，400K and 700K sentence pairs with their inter-sentence context from the LDC and combined them with the existing ZH$\rightarrow$EN training set to train the CMT-PT and DeMPT models.

Table \ref{tab:datascales} lists the performances of CMT-PT and DeMPT when extending scales of the training set into 500K (300K +200K), 700K (300k + 400K) and 1M (300K + 700K). We observe {increasing the scale of the training set consistently boosts the performance of DeMPT and CMT-PT. However, our DeMPT significantly outperforms CMT-PT across all three metrics.}

\section{Effect of Transfer Layer and Type Embedding}
\label{apdx:tsl-te-effect}
As in Eq.~\ref{equ:trans-embed} within Section \ref{sec:phase-aware}, we introduce two sublayers: a non-linear transfer sublayer and a type embedding sublayer for the trainable prompt in each phase. This design enhances the awareness of LLMs regarding the distinctions in inputs across the three tuning phases, allowing them to adapt to specific roles at each phase. We investigate the effect of these two sublayers. 

As shown in Table \ref{tab:type-trans-ctx}, our observations reveal that the transfer sublayer holds greater importance than the type embedding sublayer. Removing either the non-linear transfer sublayer (\textit{w/o} Transfer.) or the type embedding sublayer (\textit{w/o} Embed.) results in a performance drop of 0.84/0.0048/0.39 or 0.45/0.0036/0.007 in BLEU/COMET/BlonDe metrics.

\begin{table}[]
\resizebox{\columnwidth}{!}{
\begin{tabular}{lccc}
\toprule
{\bf Model} &{\bf BLEU} &{\bf COMET} &{\bf BlonDe} \\
\hline
MT-PT  &30.99 &0.8520 &49.48 \\
CMT-PT &30.82 &0.8504 &49.61  \\
\hdashline
DeMPT  &32.46 &0.8649 & 50.62 \\
~ \textit{w/o} Transfer. &  31.62  & 0.8601 &  50.23   \\
~ \textit{w/o} Embed. &  32.01  & 0.8613 &  50.55     \\
~ \textit{w/o} CTX. & 31.98  & 0.8593  &  49.89    \\
\bottomrule
\end{tabular}}
\caption{Comparison of performances of the DeMPT variants on ZH$\rightarrow$EN test set. \textit{w/o} Trans. or \textit{w/o} Embed. denotes the variant without the non-linear transfer sublayer or type embedding sublayer in Eq. \ref{equ:trans-embed}. \textit{w/o} CTX. means the inter-sentence context is not available, i.e., context-agnostic DeMPT system.}
\label{tab:type-trans-ctx}
\end{table}

\section{Effect of Hyperparameter $\lambda$}
\label{apdx:lambda}

\begin{table}[]
\resizebox{\columnwidth}{!}{
\begin{tabular}{lccc}
\toprule
{\bf Model (DeMPT)} &{\bf BLEU} &{\bf COMET} &{\bf BlonDe} \\
\hline
$\lambda_1$=1/3, $\lambda_2$=1/3, ZH$\rightarrow$EN & 32.46 & 0.8649 & 50.62 \\
$\lambda_1$=1/4, $\lambda_2$=1/3, ZH$\rightarrow$EN & 32.51 & 0.8653 & 50.31   \\
\hdashline
$\lambda_1$=1/3, $\lambda_2$=1/3, FR$\rightarrow$EN & 41.92 & 0.8790 & 58.30 \\ 
$\lambda_1$=1/4, $\lambda_2$=1/3, FR$\rightarrow$EN & 41.82 & 0.8785 & 57.92 \\
\bottomrule
\end{tabular}}
\caption{Comparison of performances of the DeMPT with different combinations of $\lambda_1$ and $\lambda_2$ on ZH$\rightarrow$EN and FR$\rightarrow$EN test sets. }
\label{tab:hyper-lambda}
\end{table}

Due to the limited computational resources, we do not perform extensive experiments to find the optimal combination of $\lambda_1$ and $\lambda_2$ for different translation tasks, simply setting them to be equal. For example, verifying each combination of $\lambda_1$ and $\lambda_2$ requires 10 experiments (5 $\times$ 2 for the number of translation directions and foundation models). Therefore, we carry out targeted experiments using a combination of $\lambda_1$ and $\lambda_2$ on ZH$\rightarrow$EN and FR$\rightarrow$EN only here.

The results are reported in Table \ref{tab:hyper-lambda}. We use a smaller value for $\lambda_1$ here and observe that the BlonDe scores are more sensitive to changes $\lambda_1$ compared to BLEU and COMET. For example, a smaller $\lambda_1$ results in -0.31 and -0.38 for ZH$\rightarrow$EN and FR$\rightarrow$EN, respectively. This sensitivity may be reasonable because $\lambda_1$ is used for adjusting the utilization of inter-sentence context.

\section{Effect of Inter-sentence Context}
\label{apdx:ctx-effect}
We implement the context-agnostic (sentence-level) DeMPT system to analyze the effect of the inter-sentence context and differences with MSP. More specifically, we replace the input of LLMs in the inter-sentence context encoding phase with the intra-sentence context. In other words, we encode the intra-sentence context twice to keep the multi-phase tuning strategy in DeMPT while making the inter-sentence context unavailable.

As shown in the last row of Table \ref{tab:type-trans-ctx} (i.e., \textit{w/o} CTX), we find that the inter-sentence context is crucial for the alleviation of discourse-related issues. The BlonDe score drops by 0.73 when the inter-sentence context is unavailable. Meanwhile, our DeMPT also significantly improves the performance of LLMs in context-agnostic MT, e.g., + 0.99 BLEU score and + 0.0073 COMET score compared to the MT-PT model.

\section{Effect of Target-side Inter-sentence Context}
\label{apdx:tgt-ctx}

\begin{table}[]
\resizebox{\columnwidth}{!}{
\begin{tabular}{lccc}
\toprule
{\bf Model} &{\bf d-BLEU} &{\bf d-COMET} &{\bf d-BlonDe} \\
\hline
MT-PT (\textit{m2o}) &34.19 &0.8216 &49.48 \\
CMT-PT (\textit{m2o}) &34.06 &0.8211 &54.68  \\
DeMPT (\textit{m2o}) &35.76 &0.8316 &55.97  \\
\hdashline
CMT-PT (\textit{m2m}) &34.13 &0.8256 &55.34  \\
\bottomrule
\end{tabular}}
\caption{Comparison of performances of the models with different translation modes, i.e., with/without target-side inter-sentence context, on ZH$\rightarrow$EN test set.}
\label{tab:type-trans-ctx}
\end{table}

To enable a fair comparison, we incorporate only the source-side inter-sentence context for the model with the concatenating strategy, i.e., the CMT-PT model in the many-to-one (\textit{m2o}) translation mode, as shown in Tables 1 and 2. To further investigate the effect of target-side inter-sentence context for the concatenating strategy, we compare the CMT-PT model in the many-to-many (\textit{m2m}) translation mode to the models in the many-to-one translation mode, for the ZH$\rightarrow$EN translation task when using the \texttt{bloomz-7b1-mt} as the foundation model.

Different from the results in Tables \ref{tab:bleu-comet} and \ref{tab:BlonDe}, we report the document-level BLEU, BlonDe, and COMET scores for all models here due to the unavailability of sentence-level alignment for many-to-many model. From the experimental results, we observe that the CMP-PT (\textit{m2m}) model outperforms the CMP-PT (\textit{m2o}) model (mostly significant in terms of the d-BlonDe metric), which demonstrates the effectiveness of the target context in addressing discourse issues. However, the CMP-PT (\textit{m2m}) model still underperforms the DeMPT model across three metrics.

\end{CJK*}

\end{document}